\documentclass[runningheads]{llncs}
\usepackage{graphicx}
\usepackage{times}
\usepackage{epsfig}
\usepackage{graphicx}
\usepackage[boxed, ruled, vlined, linesnumbered]{algorithm2e}
\usepackage{xcolor}
\usepackage{booktabs}
\usepackage{amsmath}
\usepackage{amssymb}
\usepackage{multirow}
\usepackage{float}
\usepackage{enumitem}
\usepackage{amsmath}
\newcommand{\Diff}[0]{{\rm Diff}}
\newcommand{\Reg}[0]{{\rm Reg}}

\newcommand{\reg}[0]{{\rm reg}}

\begin{document}
\title{SADIR: Shape-Aware Diffusion Models for 3D Image Reconstruction}

\author{Nivetha Jayakumar\inst{1} \and
Tonmoy Hossain\inst{2} \and Miaomiao Zhang\inst{1,2}
}
\authorrunning{N. Jayakumar et al.}

\institute{Department of Electrical and Computer Engineering, \and
Department of Computer Science,\\
School of Engineering and Applied Science, University of Virginia, VA, USA}
\maketitle             
\begin{abstract}
3D image reconstruction from a limited number of 2D images has been a long-standing challenge in computer vision and image analysis. While deep learning-based approaches have achieved impressive performance in this area, existing deep networks often fail to effectively utilize the shape structures of objects presented in images. As a result, the topology of reconstructed objects may not be well preserved, leading to the presence of artifacts such as discontinuities, holes, or mismatched connections between different parts.  In this paper, we propose a shape-aware network based on diffusion models for 3D image reconstruction, named SADIR, to address these issues. In contrast to previous methods that primarily rely on spatial correlations of image intensities for 3D reconstruction, our model leverages shape priors learned from the training data to guide the reconstruction process. To achieve this, we develop a joint learning network that simultaneously learns a mean shape under deformation models. Each reconstructed image is then considered as a deformed variant of the mean shape. We validate our model, SADIR, on both brain and cardiac magnetic resonance images (MRIs). Experimental results show that our method outperforms the baselines with lower reconstruction error and better preservation of the shape structure of objects within the images.  
\end{abstract}
\section{Introduction}

The reconstruction of 3D images from a limited number of 2D images is fundamental to various applications, including object recognition and tracking~\cite{duwek20213d}, robot navigation~\cite{zelenskii2021robot}, and statistical shape analysis for disease detection~\cite{bruse2017detecting,von2018efficient}. However, inferring the complete 3D geometry and structure of objects from one or multiple 2D images has been a long-standing ill-posed problem~\cite{lin2021artificial}. A bountiful literature has been investigated to recover the data from a missing dimension~\cite{cciccek20163d,qin2018convolutional,schlemper2017deep,waibel2023dispr}. Initial approaches to address this challenge focused on solving an inverse problem of projecting 3D information onto 2D images from geometric aspects~\cite{chung2023solving}. These solutions typically require images captured from different viewing angles using precisely calibrated cameras or medical imaging machines~\cite{chen2019self,maier2013optical}. In spite of producing a good quality of 3D reconstructions, such methods are often impractical or infeasible in many real-world scenarios.

Recent advancements have leveraged deep learning (DL) techniques to overcome the limitations posed in previous methods~\cite{cetin2023attri,goodfellow2020generative,maaloe2019biva}. Extensive research has explored various network architectures for 3D image reconstruction, including UNets~\cite{nguyen20213d}, transformers~\cite{feng2021task,korkmaz2022unsupervised}, and state-of-the-art generative diffusion models~\cite{waibel2023dispr}. These works have significantly improved the reconstruction efficiency by learning intricate mappings between stacks of 2D images and their corresponding 3D volumes. While the DL-based approaches have achieved impressive results in reconstructing detailed 3D images, they often lack explicit consideration of shape information during the learning process. Consequently, important geometric structures of objects depicted in the images may not be well preserved. This may lead to the occurrence of artifacts, such as discontinuities, holes, or mismatched connections between different parts, that break the topology of the reconstructed objects.

Motivated by recent studies highlighting the significance of shape in enhancing image analysis tasks using deep networks~\cite{chen2019learning,jiang2022one,liu2021rethinking,wang2022geo,yang2022implicitatlas}, we introduce a novel shape-aware 3D image reconstruction network called SADIR. Our methodology builds upon the foundation of diffusion models while incorporating shape learning as a key component. In contrast to previous methods that mainly rely on spatial correlations of image intensities for 3D reconstruction, our SADIR explicitly incorporates the geometric shape information aiming to preserve the topology of reconstructed images. To achieve this goal, we develop a joint deep network that simultaneously learns a shape prior (also known as a mean shape) from a given set of full 3D volumes. In particular, an atlas building network based on deformation models~\cite{wang2022geo} is employed to learn a mean shape representing the average information of training images. With the assumption that each reconstructed object is a deformed variant of the estimated mean shape, we then utilize the mean shape as a prior knowledge to guide the diffusion process of reconstructing a complete 3D image from a stack of sparse 2D slices. To evaluate the effectiveness of our proposed approach, we conduct experiments on both real brain and cardiac magnetic resonance images (MRIs). The experimental results show the superiority of SADIR over the baseline approaches, as evidenced by substantially reduced reconstruction errors. Moreover, our method successfully preserves the topology of the images during the shape-aware 3D image reconstruction process.

\section{Background: Fréchet Mean via Atlas Building}
In this section, we briefly review an unbiased atlas building algorithm~\cite{joshi2004unbiased}, a widely used technique to estimate the Fréchet mean of group-wise images. With the underlying assumption that objects in many generic classes can be described as deformed versions of an ideal template, descriptors in this class arise naturally by matching the mean (also referred as atlas) to an input image~\cite{joshi2004unbiased,wang2021bayesian,zhang2013bayesian,wu2022hybrid,zhang2016low}. The resulting transformation is then considered as a shape that reflects geometric changes.

Given a number of $N$ images $\{ \mathcal{Y}_1,\cdots,\mathcal{Y}_N\}$, the problem of atlas building is to find a mean or
template image $\mathcal{S}$ and deformation fields $\phi_1,\cdots \phi_N$ with derived initial velocity fields $v_{1}, \cdots v_{t}$ that minimize the energy function

\begin{equation}
\label{eq:lddmm}
E(\mathcal{S}, \phi_n) =\sum_{n=1}^{N} \frac{1}{\sigma^2} \text{Dist} [\mathcal{S} \circ \phi_n(v_t), \mathcal{Y}_n] \, +\, \text{Reg} [\phi_n(v_t)],
\end{equation}
where $\sigma^2$ is a noise variance and $\circ$ denotes an interpolation operator that deforms image $\mathcal{Y}_n$ with an estimated transformation $\phi_n$. The $\text{Dist}[\cdot, \cdot]$ is a distance function that measures the dissimilarity between images, i.e., sum-of-squared differences~\cite{beg2005computing}, normalized cross correlation~\cite{avants2008symmetric}, and mutual information~\cite{wells1996multi}. The $\Reg[\cdot]$ is a regularizer that guarantees the smoothness of transformations.

Given an open and bounded $d$-dimensional domain $\mathrm{\Omega} \subset \mathbb{R}^d$, we use $\Diff(\mathrm{\Omega})$ to denote a space of diffeomorphisms (i.e., a one-to-one smooth and invertible smooth transformation) and its tangent space $V=T\Diff(\mathrm{\Omega})$. A well-developed algorithm, large deformation diffeomorphic metric mapping (LDDMM)~\cite{beg2005computing}, provides a regularization that guarantees the smoothness of deformation fields and preserves the topological structures of objects for the atlas building framework (Eq.~\eqref{eq:lddmm}). Such a regularization is formulated as an integral of the Sobolev norm of the time-dependent velocity field $v_n(t) \in V(t \in [0,1])$ in the tangent space, i.e.,
\begin{align}
    &\Reg[\phi_n(v_t)] = \int_0^1 (L v_t, v_t) \, dt,   \quad \text{with} \quad \frac{d\phi_n(t)}{dt} = - D\phi_n(t)\cdot v_n(t),   
    \label{eq:distance}
\end{align}
where $L: V\rightarrow V^{*}$ is a symmetric, positive-definite differential operator that maps a tangent vector $ v_t\in V$ into its dual space as 
a momentum vector $m_t \in V^*$. We write $m_t = L v_t$, or $v_t = K m_t$, with $K$ being an inverse operator of $L$. The operator $D$ denotes a Jacobian matrix and $\cdot$ represents element-wise matrix multiplication. In this paper, we use a metric of the form $L=(-\alpha \Delta + \gamma \mathbf{I})^3$, in which $\Delta$ is the discrete Laplacian operator, $\alpha$ is a positive regularity parameter that controls the smoothness of transformation fields, $\gamma$ is a weighting parameter, and $\mathbf{I}$ denotes an identity matrix.

The minimum of Eq.~\eqref{eq:distance} is uniquely determined by solving an Euler-Poincar\'{e} differential equation (EPDiff)~\cite{arnold1966geometrie,miller2006geodesic} with a given initial condition of velocity fields, noted as $v_0$. This is known as the {\em geodesic shooting} algorithm~\cite{vialard2012diffeomorphic}, which nicely proves that the deformation-based shape descriptor $\phi_n$ can be fully characterized by an initial velocity field $v_n(0)$. The mathmatical formulation of the EPDiff equation is
\begin{align}
    \frac{\partial v_n(t)}{\partial t} = - K \left[(D v_n(t))^T \cdot m_n(t) + D m_n(t) \cdot v_n(t) + m_n(t) \cdot \operatorname{div} v_n(t) \right],
    \label{eq:epdiff}
\end{align}
where the operator $D$ denotes a Jacobian matrix, $\operatorname{div}$ is the divergence, and $\cdot$ represents element-wise matrix multiplication.

We are now able to equivalently minimize the atlas building energy function in Eq.~\eqref{eq:lddmm} as
\begin{equation}
   E(\mathcal{S}, \phi_n) =  \sum_{n=1}^{N} \frac{1}{\sigma^2} \text{Dist} [\mathcal{S} \circ \phi_n(v_n(t)), \mathcal{Y}_n] + (L v_n(0), v_n(0)), \, \,  \text{s.t. Eq.~\eqref{eq:distance} \&~\eqref{eq:epdiff}.}   \label{eq:flddmm}
\end{equation}
For notation simplicity, we will drop the time index in the following sections. 

\section{Our Method: SADIR}
In this section, we present SADIR, a novel reconstruction network that incorporates shape information in predicting 3D volumes from a limited number of input 2D images. We introduce a sub-module of the atlas building framework, which enables us to learn shape priors from a given set of full 3D images. It is worth mentioning that while the backbone of our proposed SADIR is a diffusion model~\cite{ho2020denoising}, the methodology can be generalized to a variety of network architectures such as UNet~\cite{ronneberger2015u}, UNet++~\cite{zhou2018unet++}, and Transformer~\cite{dosovitskiy2020image}.

\subsection{Shape-Aware Diffusion Models Based on Atlas Building Network}
Given a number of $N$ training data $\{I_n, \mathcal{Y}_n\}_{n=1}^N$, where $I_n$ is a stack of sparse 2D images with its associated full 3D volume $\mathcal{Y}_n$. Our model SADIR consists of two submodules: 
\begin{enumerate}[label=(\roman*)]
\item An atlas building network, parameterized by $\theta^a$, that provides a mean image $\mathcal{S}$ of $\{\mathcal{Y}_n\}$. In this paper, we employ the network architecture of Geo-SIC~\cite{wang2022geo};
\item A reconstruction network, parameterized by $\theta^r$, that considers each reconstructed image $\hat{\mathcal{Y}}_n$ as a deformed variant of the obtained atlas, i.e., $\hat{\mathcal{Y}}_n \overset{\Delta}{=} \mathcal{S} \circ \phi_n(v_n(\theta^r))$. In contrast to current approaches learning the reconstruction process based on image intensities, our model is developed to learn the geometric shape variations represented by the predicted velocity field $v_n$. 
\end{enumerate}

Next, we introduce the details of our shape-aware diffusion models for reconstruction, which is a key component of SADIR. Similar to existing diffusion models~\cite{ho2020denoising,waibel2023dispr}, we develop a forward diffusion and a reverse diffusion process to predict the velocity fields associated with the pair of input training images and an atlas image. For the purpose of simplified math notations, we omit the index $n$ for each subject in the following sections. \\ 

\noindent \textbf{Forward diffusion process.} Let $y^0$ denote the original 3D image with full volumes and $\tau$ denote the time point of the diffusion process. We assume the data distribution of $y^\tau$ is a normal distribution with mean $\mu$ and variance $\beta$, i.e., $y^\tau \sim \mathcal{N}(\mu, \beta)$. The forward diffusion of $y^{\tau-1}$ to $y^{\tau}$ is then recursively given by
\begin{equation}
    p(y^\tau \, | \, y^{\tau-1}) = \mathcal{N}(y^\tau; \sqrt{1-\beta^\tau}y^{\tau-1}, \beta^\tau \mathbf{I}),
    \label{eq:forward_diff}
\end{equation}
where $\mathbf{I}$ denotes an identity matrix, and $\beta^\tau \in [0, 1]$ denotes a known variance increased along the time steps with $\beta^1 < \beta^2 < \cdots < \beta^\tau$. The forward diffusion process
is repeated for a fixed, predefined number of time steps. 

It is shown in~\cite{ho2020denoising} that repeated application of Eq.~\eqref{eq:forward_diff} to the original image $y^0$ and setting $\alpha^\tau = 1 - \beta^\tau$ and $\Bar{\alpha}^\tau =\prod^{\tau}_{i=1}\alpha^i$ yields
\begin{align*}
 p(y^\tau \, | \, y^0) = \mathcal{N}(y^\tau; \sqrt{\Bar{\alpha}^\tau}y^0, (1-\Bar{\alpha}^\tau)\mathbf{I}). 
\end{align*}

Therefore, we can write $y^\tau$ in terms of $y^0$ as
\begin{align*}
    y^\tau = \sqrt{\Bar{\alpha}^\tau}y^0 + \sqrt{1-\Bar{\alpha}}^\tau\epsilon \quad \mathrm{with} \quad \epsilon \sim \mathcal{N}(0,\mathbf{I}).
\end{align*}

\noindent \textbf{Reverse diffusion process.} Given a concatenation of a sparse stack of 2D images $I$, an atlas image $\mathcal{S}$, and $y^\tau$ from the forward process, our diffusion model is designed to remove the added noise in the reverse process. Following the work of~\cite{wolleb2022diffusion}, we will now predict $y^{\tau-1}$ from the input $y^\tau$. The joint probability distribution $p(y^{\tau-1} \, | \, y^\tau)$ is predicted by a trained neural network (e.g., UNet) in each reverse time step for all $\tau \in \{1, \cdots, T\}$, where $T$ is the maximal time step. With the network model parameters denoted by $\theta^r$, we can write the reverse process as
\begin{align*}
     p_{\theta^r}(y^{\tau-1} \, | \, y^{\tau}) = \mathcal{N}(y^{\tau-1};\mu_{\theta^r}(y^{\tau}, \tau),\mathrm{\Sigma}_{\theta^r}(y^{\tau}, \tau)). 
\end{align*}

Similarly, we can write $y^{\tau-1}$ backward in terms of $y^\tau$ as
\begin{align*}
     y^{\tau-1} = \frac{1}{\sqrt{\alpha^\tau}}(y^\tau\frac{1-\alpha^\tau}{\sqrt{1-\Bar{\alpha}^\tau}}\epsilon_{\theta^r}(y^\tau,\tau)) + \sigma^t\mathbf{z},
\end{align*}
where $\sigma^\tau$ is the variance scheme the model can learn, the component $\mathbf{z}$ is a stochastic sampling process. The
model is trained with input $y^\tau$ to subtract the
noise scheme $\epsilon_{\theta^r}(y^\tau,\tau)$ from $y^\tau$ to produce $y^{\tau-1}$.

The output of this reverse process is a predicted velocity field $v(\theta^r)$, which is then used to generate its associated transformation $\phi (v(\theta^r))$ to deform the atlas $S$. Such a deformed atlas is the reconstructed image $\hat{\mathcal{Y}} = \mathcal{S} \circ \phi (v(\theta^r))$. 

\begin{figure}[h]
    \centering
    \includegraphics[width=\textwidth]{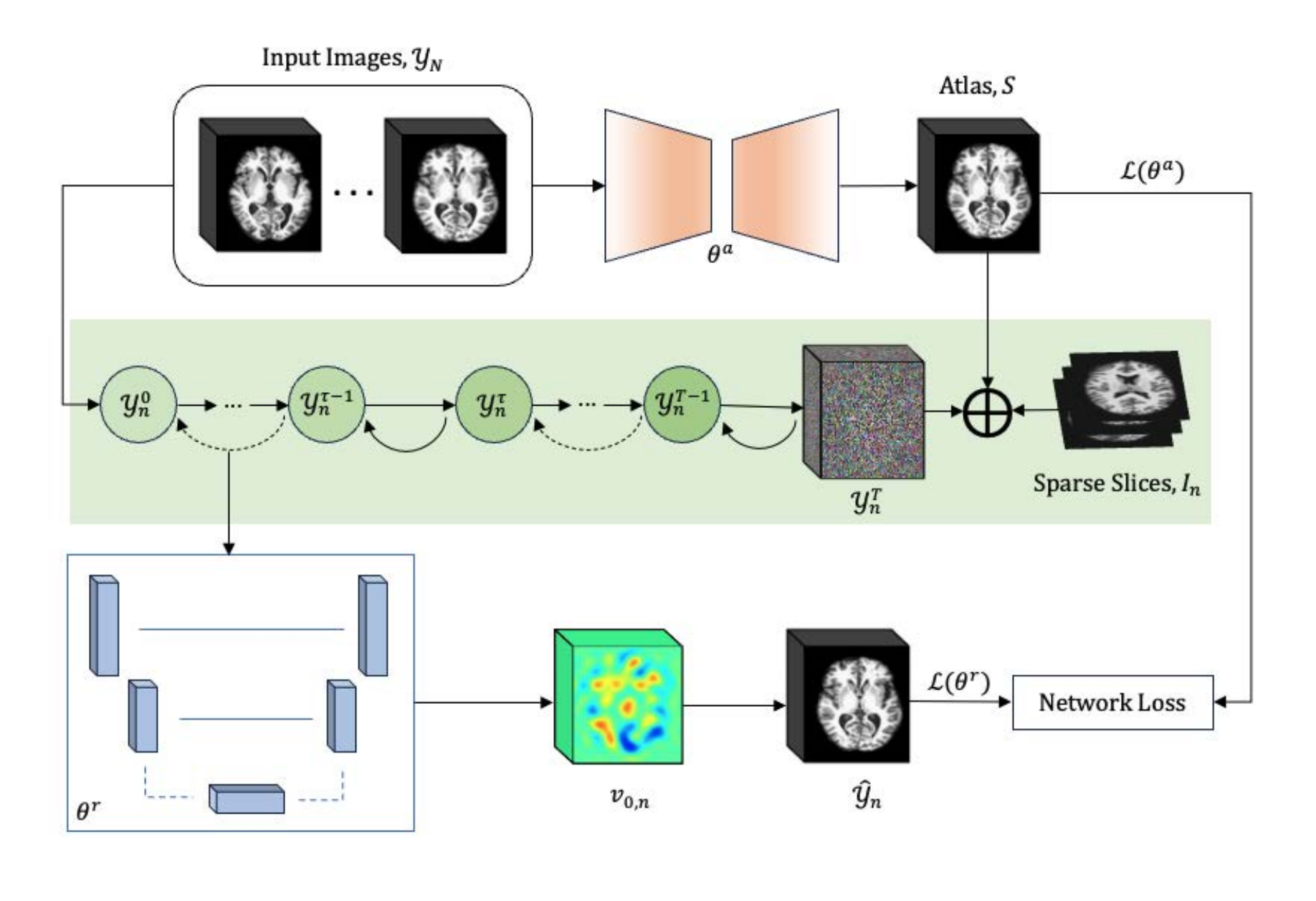}
    \caption{An overview of our proposed 3D reconstruction model SADIR.}
    \label{fig:model}
\end{figure}

An overview of the proposed SADIR network architecture is shown in Fig.~\ref{fig:model}.

\subsection{Network Loss and Optimization}
The network loss function of our model, SADIR, is a joint loss of the atlas building network and the diffusion reconstruction network. We first define the atlas building loss as   
 \begin{equation}
     \mathcal{L}(\theta^a) = \sum_{n=1}^{N} \frac{1}{\sigma^2} \| \mathcal{S}(\theta^a) \circ (\phi_n(v_n)) - \mathcal{Y}_n \|_2^2 + (L v_n, v_n) + \reg(\theta^a),
     \label{eq:atl_bd}
\end{equation}
where $\reg(\cdot)$ denotes a regularization on the network paramters.   

We then define the loss function of the diffusion reconstruction network as a combination of sum-of-squared differences and S\o rensen$-$Dice coefficient~\cite{dice1945measures} loss (for distinct anatomical structure, e.g., brain ventricles or myocardium) between the predicted reconstruction and ground-truth in following
\begin{equation}
    \mathcal{L}(\theta^r) = \sum_{n=1}^{N}\|  \mathcal{S} \circ\phi_n(v_n(\theta^r)) - \mathcal{Y}_n \|^2_2 + \eta\, [1-\text{Dice} (\mathcal{S} \circ\phi_n(v_n(\theta^r)),\mathcal{Y}_n)] + \reg(\theta^r),
    \label{eq:diffusion}
\end{equation}

where $\eta$ is the weighting parameter, and $\text{Dice}(\hat{\mathcal{Y}},\mathcal{Y}_n) = 2(|\hat{\mathcal{Y}}| \cap |\mathcal{Y}_n|)/(|\hat{\mathcal{Y}}| + |\mathcal{Y}_n|)$, considering $\hat{\mathcal{Y}}_n \overset{\Delta}{=} \mathcal{S} \circ \phi_n(v_n(\theta^r))$. Defining $\lambda$ as a weighting parameter, we are now ready to write the joint loss of SADIR as 
\begin{equation*}
     \mathcal{L} = 
     \mathcal{L}(\theta^a) + \lambda \mathcal{L}(\theta^r).
     \label{eq:tloss}
 \end{equation*}

\noindent {\bf Joint network learning with an alternative optimization.} We use an alternative optimization scheme~\cite{nocedal1999numerical} to minimize the total loss $\mathcal{L}$ in Eq.~\eqref{eq:tloss}. More specifically, we jointly optimize all network parameters by alternating between the training of the atlas building and diffusion reconstruction network, making it end-to-end learning. A summary of our joint training of SADIR is presented in Alg.~\ref{alg:altopt}.

\begin{algorithm}[!ht]
\SetAlgoLined
\SetArgSty{textnormal}
\SetKwInOut{Input}{Input}
\SetKwInOut{Output}{Output}
  \Input{A group of $N$ input images with full 3D volumes $\{\mathcal{Y}_n\}$ and a stack of sparse 2D images $\{I_n\}$.}
  \Output{Generate mean shape or atlas $\mathcal{S}$, initial velocity fields $v_n$, and reconstructed images $\hat{\mathcal{Y}}_n$}
  
   \For{i = 1 to $p$}
   {

   \tcc{Train geometric shape learning network}
   
   Minimize the atlas building loss in Eq.~\eqref{eq:atl_bd}
   
   Output the atlas $\mathcal{S}$
   
   \tcc{Train diffusion network}
   
   Minimize the diffusion reconstruction loss in Eq.~\eqref{eq:diffusion}
   
   Output the initial velocity fields $\{v_{n}\}$ 
   and the reconstructed images $\hat{\mathcal{Y}}_n$
   }
   \textbf{Until convergence}
\caption{Joint Training of SADIR.}
\label{alg:altopt}
\end{algorithm}

\section{Experimental Evaluation}
We demonstrate the effectiveness of our proposed model, SADIR, for 3D image reconstruction from 2D slices on both brain and cardiac MRI scans. \\

\noindent \textbf{3D Brain MRIs:} For 3D real brain MRI scans, we include $214$ public T1-weighted longitudinal brain scans from the latest released Open Access Series of Imaging Studies (OASIS-III) \cite{lamontagne2019oasis}. All subjects include both healthy and disease individuals, aged from $42$ to $95$. All MRIs were pre-processed as $256 \times 256 \times 256$, $1.25mm^{3}$ isotropic voxels, and underwent skull-stripped, intensity normalized, bias field corrected and pre-aligned with affine transformation. To further validate the performance of our proposed model on specific anatomical shapes, we select left and right brain ventricles available in the OASIS-III dataset~\cite{lamontagne2019oasis}.\\

\noindent \textbf{3D Cardiac MRIs:} For 3D real cardiac MRI, we include $215$ publicly available 3D myocardium mesh data from MedShapeNet dataset~\cite{Li2023}. We convert the mesh data to binary label maps using 3D slicer~\cite{Fedorov3dslicer}. All the images were pre-processed as $222 \times 222 \times 222$ and pre-aligned with affine transformation.

\subsection{Experimental Settings}
We first validate our proposed model, SADIR, on reconstructing 3D brain ventricles, as well as brain MRIs from a sparse stack of eight 2D slices. We compare our model's performance with three state-of-the-art deep learning-based reconstruction models: 3D-UNet~\cite{cciccek20163d}; DDPM, a probabilistic diffusion model~\cite{ho2020denoising}; and DISPR, a diffusion model based shape reconstruction model with geometric topology considered~\cite{waibel2023dispr}. Three evaluation metrics, including the Sørensen–Dice coefficient (DSC)~\cite{dice1945measures}, Jaccard Similarity~\cite{jaccard1908nouvelles}, and RHD95 score~\cite{rhd95}, are used to validate the prediction accuracy of brain ventricles for all methods. For brain MR images, we show the error maps of reconstructed images for all the experiments.

To further validate the performance of SADIR on different datasets, we run tests on a relatively small dataset of cardiac MRIs to reconstruct 3D myocardium. \\

\noindent \textbf{Parameter setting:} We set the mean and standard deviation of the forward diffusion process to be $0$ and $0.1$, respectively. The scheduling is $\operatorname{linear}$ for the noising process and is scaled to reach an isotropic Gaussian distribution irrespective of the value of $T$. For the atlas building network, we set the depth of the UNet architecture as $4$. We set the number of time steps for Euler integration in EPDiff (Eq.~\eqref{eq:epdiff}) as $10$, and the noise variance $\sigma=0.02$. For the shooting, we use a kernel map valued $[0.5, 0, 1.0]$. Besides, we set the parameter $\alpha=3$ for the operator $L$. Similar to \cite{waibel2023dispr}, we set the batch size as $1$ for all experiments. We utilize the cosine annealing learning rate scheduler that starts with a learning rate of $\eta=1\text{e}^{-3}$ for network training. We run all models on training and validation images using the Adam optimizer and save the networks with the best validation performance.

 In the reverse process of the diffusion network, we set the depth of the 3D attention-UNet backbone as 6. We introduce the attention mechanism via spatial excitation channels \cite{hu2019squeezeandexcitation}, with ReLU (Rectified Linear Unit) activation. The UNet backbone has ELU activation (Exponential Linear Unit) in the hidden convolution layers and GeLU (Gaussian error Linear Unit) activation with tanh approximation. For each training experiment, we utilize Rivanna (high-performance computing servers of the University of Virginia) with NVIDIA A100 and V100 GPUs for $\sim18$ hours (till convergence). For all the experimental datasets, we split all the training datasets into $70\%$ training, $15\%$ validation, and $15\%$ testing. For both training and testing, we downsample all the image resolutions to $64 \times 64 \times 64$. 

\subsection{Experimental Results}
\begin{figure}[h]
    \centering
    \includegraphics[width=0.93\textwidth]{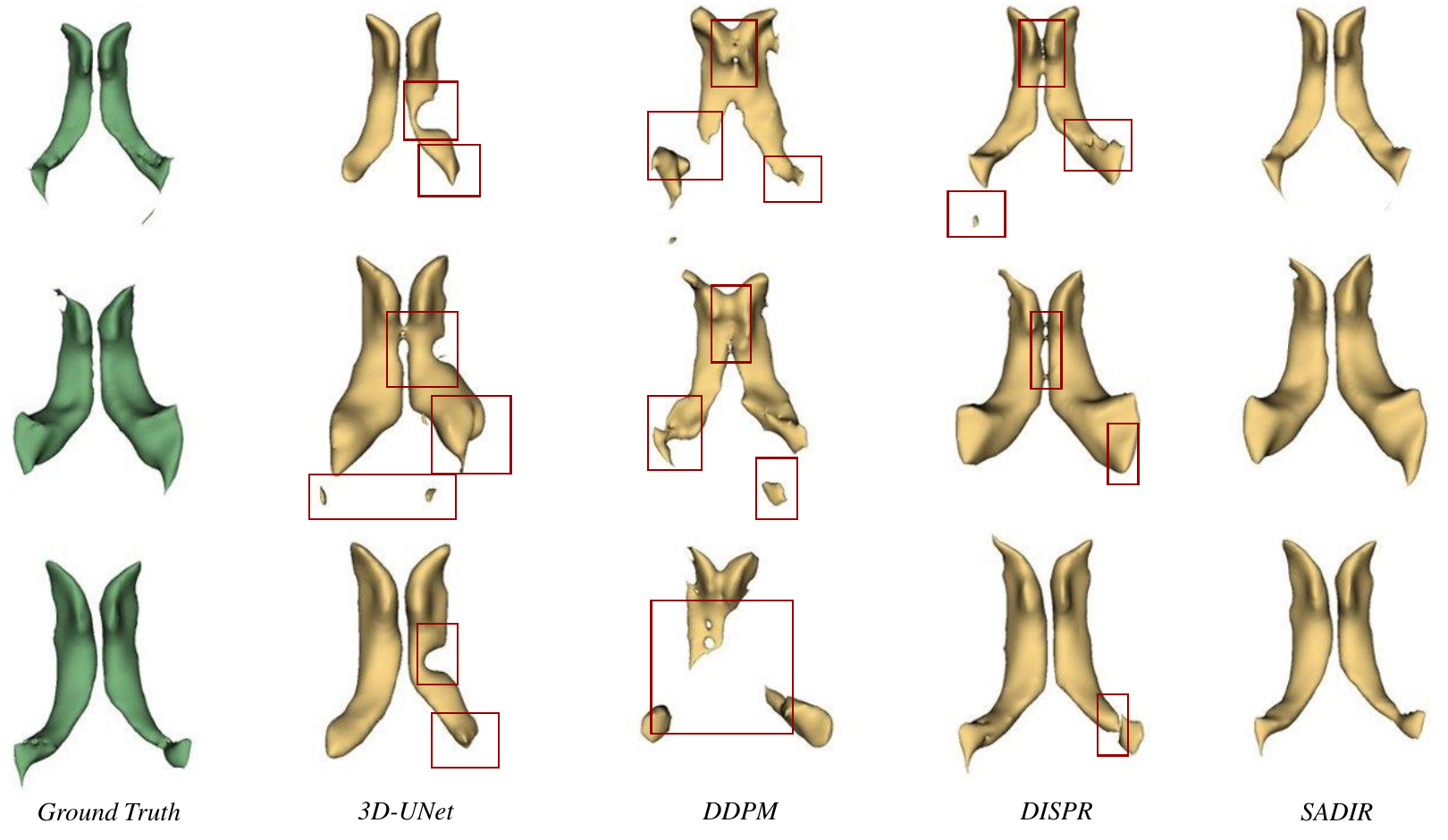}
    \caption{Top to bottom: examples of reconstructed 3D brain ventricles from sparse 2D slices; Left to right: a comparison of brain ventricles of all reconstruction models with ground truth.}
    \label{fig:ventricles}
\end{figure}

Fig.~\ref{fig:ventricles} visualizes examples of ground truth and reconstructed 3D volumes of brain ventricles from all methods. It shows that SADIR outperforms all baselines in well preserving the structural information of the brain ventricles. In particular, models without considering the shape information of the images (i.e., 3D-UNet and DDPM) generate unrealistic shapes such as those with joint ventricles, holes in the volume, and deformed ventricle tails. While the other algorithm, DISPR, shows improved performance of enforcing topological consistency on the object surface, its predicted results of 3D volumes are inferior to SADIR. 

Tab.~\ref{tab:vent} reports the average scores along with the standard deviation of the Dice similarity coefficient (DSC), Jaccard similarity, and Hausdorff distance computed between the brain ventricles reconstructed by all the models and the ground truth. Compared to all the baselines, SADIR achieves the best performance with a $1.6\%-5.6\%$ increase in the average DSC with the lowest standard deviations across all metrics.

\begin{table}[htbp]
\centering
\caption{A comparison of 3D brain ventricle reconstruction for all methods.}
\begin{tabular}{lccc}
\toprule
Model & DSC $\uparrow$ & Jaccard similarity $\uparrow$ & RHD95 $\downarrow$ \\
\midrule
3D-Unet & 0.878 $\pm$ 0.0128 & 0.804 $\pm$ 0.0204 & 4.366 $\pm$ 1.908 \\ 
DDPM\quad\quad & 0.731 $\pm$ 0.0292 & 0.652 $\pm$ 0.0365 & 8.827 $\pm$ 9.212\\
DISPR & 0.918 $\pm$ 0.0097& 0.861 $\pm$ 0.0158& \textbf{1.041 $\pm$ 0.130}\\
\textbf{SADIR} & \textbf{0.934 $\pm$ 0.013} & \textbf{0.900 $\pm$ 0.021} & 1.414 $\pm$ 0.190 \\
\bottomrule
\end{tabular}
\label{tab:vent}
\end{table}

\begin{figure}[!h]
    \centering
    \includegraphics[width=0.95\textwidth]{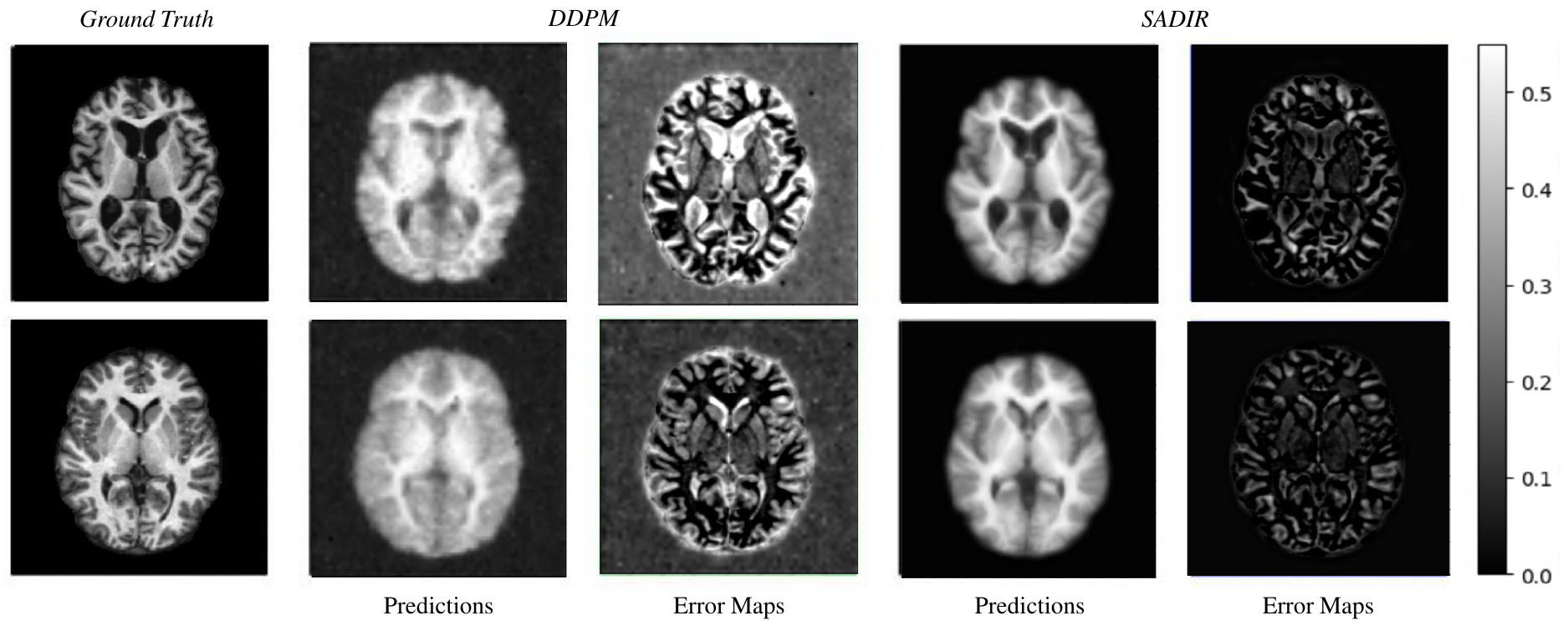}
    \caption{Left to right: a comparison of ground truth, DDPM, and SADIR along with the error map.}
    \label{fig:cmap}
\end{figure}

Fig. \ref{fig:cmap} visualizes the ground truth and reconstructed 3D brain MRIs as a result of evaluating DDMP and our method SADIR on the test data, along with their corresponding error maps. The error map is computed as absolute values of an element-wise subtraction between the ground truth and the reconstructed image. The images reconstructed by SADIR outperform the DDPM with a low absolute reconstruction error. Our method also preserves crucial anatomical features such as the shape of the ventricles, corpus callosum and gyri, which cannot be seen in the images reconstructed by the DDPM. This can be attributed to the lack of incorporating the shape information to guide the 3D MRI reconstruction. Moreover, our model has little to no noise in the background as compared to the DDPM.  

Tab.~\ref{tab:myo} reports the average scores of DSC, Jaccard similarity, and Hausdorff distance evaluated between the reconstructed myocardium from all algorithms and the ground truth. Our method proves to be competent in reconstructing 3D volumes without discontinuities, artifacts, jagged edges or amplified structures, as can be seen in results from the other models. Compared to the baselines, SADIR achieves the best performance in terms of DSC, Jaccard similarity, and RHD95 with the lowest standard deviations across all metrics.

\begin{table}[H]
\centering
\caption{A comparison of 3D myocardium reconstruction for all methods.}
\begin{tabular}{lccc}
\toprule
Model & DSC $\uparrow$ & Jaccard similarity $\uparrow$ & RHD95 $\downarrow$\\
\midrule
3D-Unet & 0.870 $\pm$ 0.0158 & 0.771 $\pm$ 0.024 & 0.840 $\pm$ 0.202 \\ 
DDPM\quad\quad & 0.823 $\pm$ 0.014 & 0.668 $\pm$ 0.019 & 1.027 $\pm$ 0.093\\
DISPR & 0.950 $\pm$ 0.017 & 0.906 $\pm$ 0.031 & 0.347 $\pm$ 0.032\\
\textbf{SADIR} & \textbf{0.978 $\pm$ 0.016} & \textbf{0.957 $\pm$ 0.031} & \textbf{0.341 $\pm$ 0.023} \\
\bottomrule
\end{tabular}
\label{tab:myo}
\end{table}

\begin{figure}[!h]
    \centering
    \includegraphics[width=0.85\textwidth]{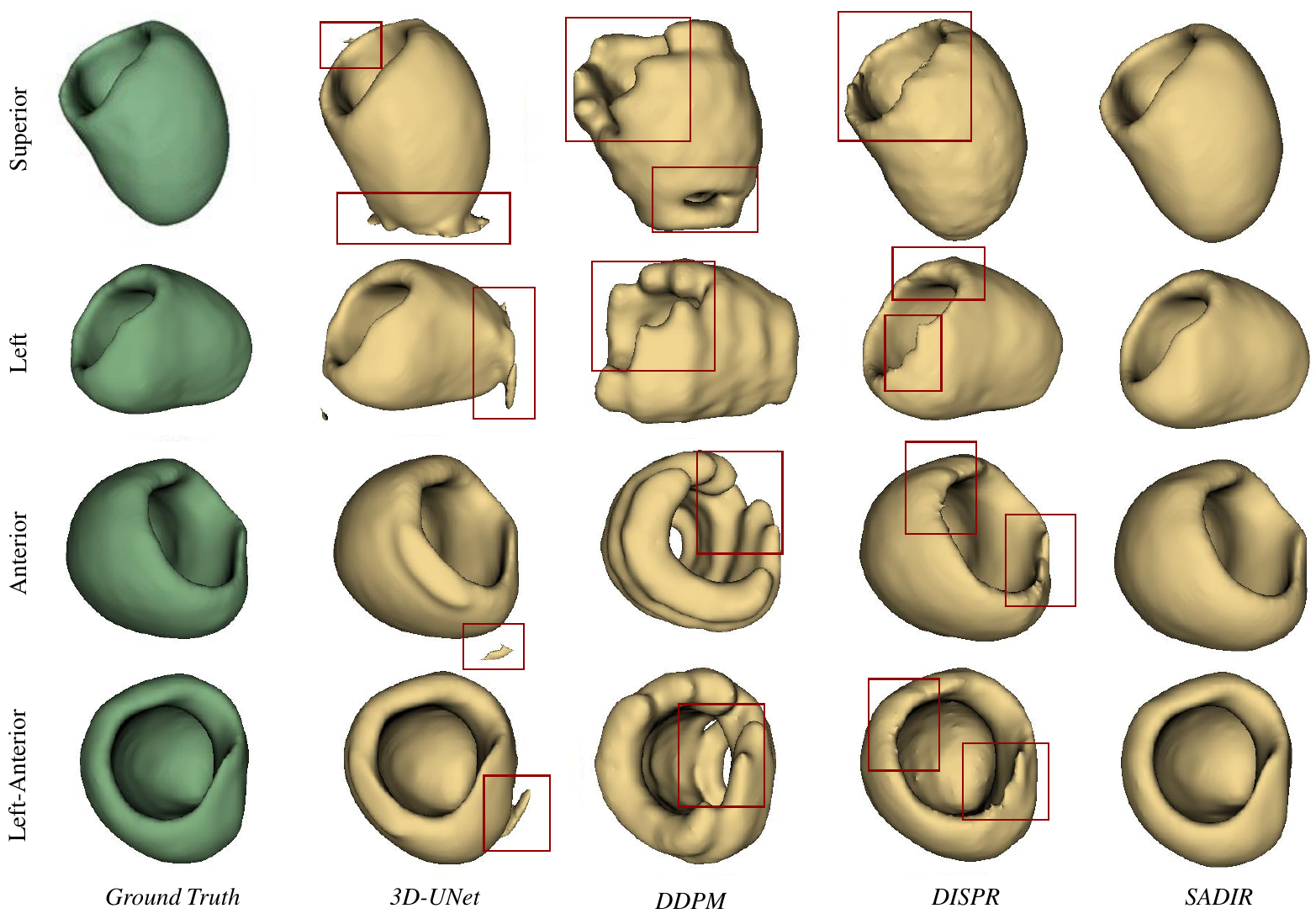}
    \caption{A comparison of reconstructed 3D myocardium between ground truth, 3D-UNet, DDPM, DISPR, and SADIR over four different views.}
    \label{fig:myocomp}
\end{figure}

Fig.~\ref{fig:myocomp} visualizes a comparison of the reconstructed 3D myocardium between the ground truth and all models. It shows that our method consistently produces reconstructed volumes that preserve the original shape of the organ with less artifacts. 

\begin{figure}[H]
    \centering
    \includegraphics[width=0.7\textwidth]{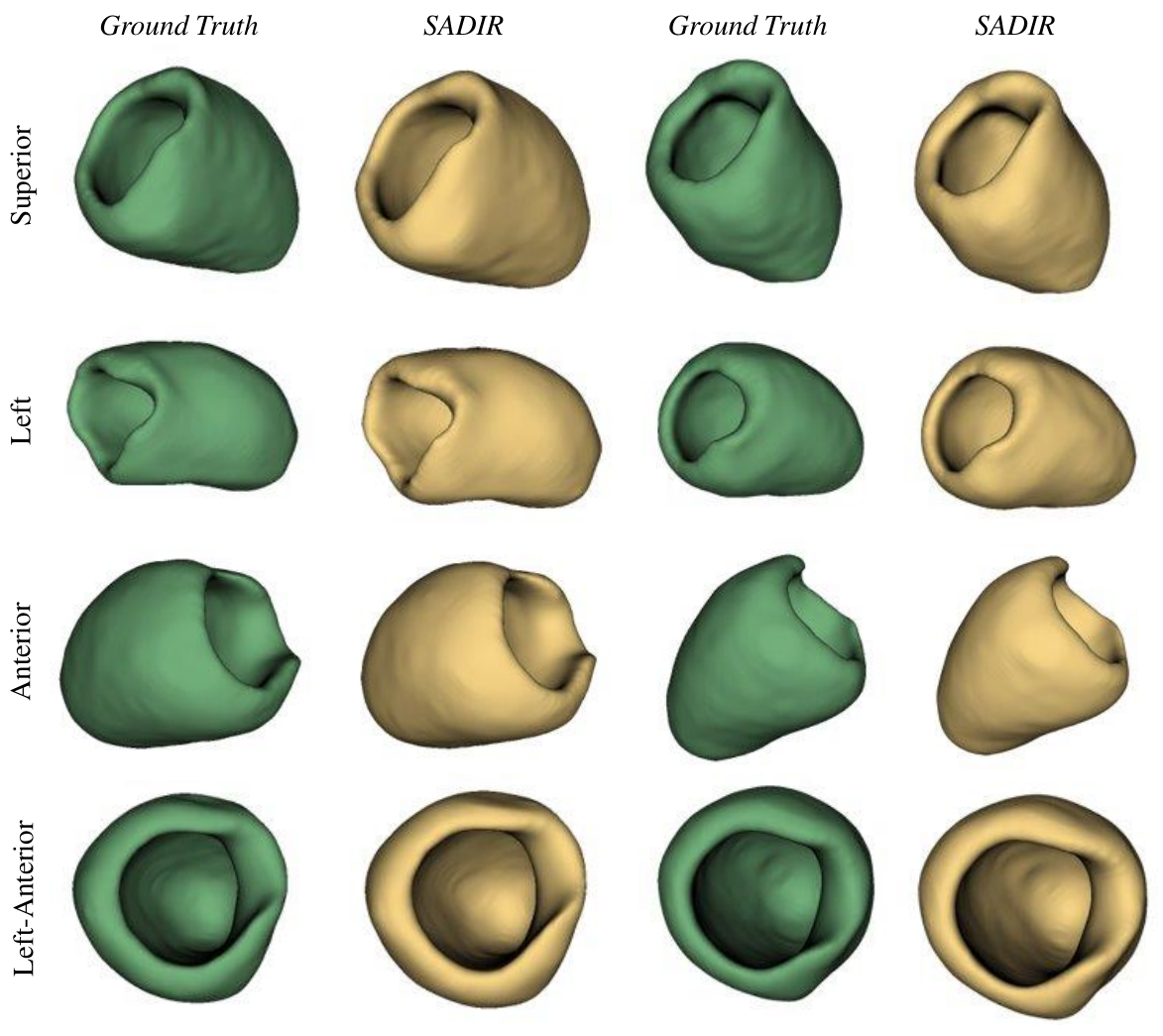}
    \caption{3D myocardium reconstructed from sparse 2D slices by SADIR over four different views.}
    \label{fig:myo}
\end{figure}

Fig.~\ref{fig:myo} shows examples of the superior, left, anterior and left-anterior views of the 3D ground truth and SADIR-reconstructed volumes of the myocardium for different subjects. We observe that the results predicted by SADIR have little to no difference from the ground truth, thereby efficiently preserving the anatomical structure of the myocardium.

\section{Conclusion}
This paper introduces a novel shape-aware image reconstruction framework based on diffusion model, named as SADIR. In contrast to previous approaches that mainly rely on the information of image intensities, our model SADIR incorporates shape features in the deformation spaces to preserve the geometric structures of objects in the reconstruction process. To achieve this, we develop a joint deep network that simultaneously learns the underlying shape representations from the training images and utilize it as a prior knowledge to guide the reconstruction network. To the best of our knowledge, we are the first to consider deformable shape features into the diffusion model for the task of image reconstruction. Experimental results on both 3D brain and cardiac MRI show that our model efficiently produces 3D volumes from a limited number of 2D slices with substantially low reconstruction errors while better preserving the topological structures and shapes of the objects.

\paragraph*{\bf Acknowledgement.} This work was supported by NSF CAREER Grant 2239977 and NIH 1R21EB032597.

\bibliographystyle{abbrv}
\bibliography{references}
\end{document}